\documentclass[conference]{IEEEtran}
\IEEEoverridecommandlockouts
\usepackage{cite}
\usepackage{amsmath,amssymb,amsfonts}
\usepackage{algorithmic}
\usepackage{graphicx}
\usepackage{textcomp}
\usepackage{xcolor}
\usepackage{booktabs}
\usepackage{subcaption}
\def\BibTeX{{\rm B\kern-.05em{\sc i\kern-.025em b}\kern-.08em
    T\kern-.1667em\lower.7ex\hbox{E}\kern-.125emX}}
\begin{document}

\title{Human Fall Detection- Multimodality Approach}

\author{
\IEEEauthorblockN{Xi Wang}
\IEEEauthorblockA{\textit{Multimedia} \\
\textit{University of Alberta}\\
Edmonton, Canada \\
xi29@ualberta.ca}
\and
\IEEEauthorblockN{Ramya Penta}
\IEEEauthorblockA{\textit{Multimedia} \\
\textit{University of Alberta}\\
Edmonton, Canada \\
penta@ualberta.ca}
\and
\IEEEauthorblockN{Bhavya Sehgal}
\IEEEauthorblockA{\textit{Multimedia} \\
\textit{University of Alberta}\\
Edmonton, Canada \\
bsehgal1@ualberta.ca}
\and
\IEEEauthorblockN{Dale Chen-Song}
\IEEEauthorblockA{\textit{Multimedia} \\
\textit{University of Alberta}\\
Edmonton, Canada \\
chensong@ualberta.ca}
}

\maketitle

\begin{abstract}
Falls have become more frequent in recent years, which has been harmful for senior citizens.Therefore detecting falls have become important and several data sets and machine learning model have been introduced related to fall detection. In this project report, a human fall detection method is proposed using a multi modality approach. We used the UP-FALL detection data set which is collected by dozens of volunteers using different sensors and two cameras. We use wrist sensor with acclerometer data keeping labels to binary classification, namely fall and no fall from the data set.We used fusion of camera and sensor data to increase performance. The experimental results shows that using only wrist data as compared to multi sensor for binary classification did not impact the model prediction performance for fall detection. 

\end{abstract}


\section{Introduction}
Falling is a serious problem that can occur for elderly people as it can cause serious injuries, such as fractured hips, traumatic brain injuries, and other injuries \cite{spineinjuries2007}. These injuries can be exasperated if the elderly people are living alone without the help of a caretaker, as they can stay on the ground for an extended period of time unable to get up, losing consciousness, getting hypothermia, and may even lead to death \cite{tinetti1993predictors}. Therefore, immediate observations for fall detection of elderly people using fall detection devices are necessary to prevent such injuries.

There are different systems of monitoring and detecting human falling on a daily basis. There are two types of classification these systems can fall under wearable devices and non-wearable devices. Wearable devices include devices that can be attached to the person, such as accelerometers \cite{bagala2012evaluation}, gyroscopes \cite{huynh2015optimization}, and wearable cameras \cite{ozcan2013automatic} that can monitor fall actions. Some disadvantages of these types of devices are that they may be cumbersome and uncomfortable to constantly wear the device and they may need to be recharged consistently. Furthermore, the devices can be even more uncomfortable if the subject needs to wear more than one. Non-wearable devices can include optical cameras \cite{auvinet2008fall}, microphones \cite{li2012microphone}, microwave radar \cite{liu2011automatic}, or depth cameras \cite{bian2014fall}. However, these also have limitations such as cameras being sensitive to light intensity, microphones needing a quiet environment to detect sounds accurately, and microwave characteristics of human action are not robust enough due to the range and Doppler resolution limits of radar system \cite{zhou2018fall}. Therefore, the best way to mitigate these disadvantages is to combine different systems using a multimodality approach. However, another challenge comes in determining which systems should be combined for the multimodality approach, as some systems may work better with others or be redundant.

Furthermore, there is another challenge to fall detection; the subject needs to be confirmed falling. There are certain movements or actions that the devices may detect that may not be actually falling. Therefore, the devices would need to distinguish between common actions such as walking, laying down, and standing up, and common falling false alarms such as squatting or picking up stuff from the ground and then the actual falling itself. 

In this report, we will propose a multimodal method to detect human falling from camera and accelerometer data, using feature fusion and multi-layer perceptron.

\section{Background}
\subsection{Multimodality Approach}
As mentioned above, there are many devices that can be used to detect falling. Therefore, there are many ways to combine these systems for a multimodality approach. \cite{multimodaldeeplearning} proposes a novel application of deep networks to learn features over multiple modalities. They presented a series of tasks for multimodal learning and show how to train deep networks that learn features to address these tasks. They demonstrated cross-modality feature learning, where better features for one modality (e.g., video) can be learned if multiple modalities (e.g., audio and video) are present at feature learning time. Furthermore, they show how to learn a shared representation between modalities and evaluate it on a unique task, where the classifier is trained with audio-only data but tested with video-only data and vice-versa.

The overall task was divided into three phases: feature learning, supervised training, and testing. A simple linear classifier was used for supervised training and testing to examine different feature learning models with multimodal data. In particular, they considered three learning settings: multimodal fusion, cross-modality learning, and shared representation learning.

One of the most straightforward approaches to feature learning is to train a Restricted Boltzmann Machines (RBMs) model separately for audio and video but it had two issues with it. First, there was no explicit objective for the models to discover correlations across the modalities; it is possible for the model to find representations such that some hidden units are tuned only for audio while others are tuned only for video. Second, the models are clumsy to use in a cross-modality learning setting where only one modality is present during supervised training and testing. 

Thus, they proposed a deep autoencoder that resolves both issues. cross-modality learning setting was used where both modalities are present during feature learning but only a single modality is used for supervised training and testing. They used the deep autoencoder models in settings where only a single modality is present at supervised training and testing. They proposed training the bimodal deep autoencoder using an augmented but noisy dataset with additional examples that have only a single modality as input. Both models are pre-trained using sparse RBMs.

\cite{cnnwu2022} In EO, which is earth observation tasks, using single remote sensor data like hyperspectral data or Lidar data, researchers inevitably meet the performance bottleneck in identifying and recognizing objects of interest due to the difficulties in excavating and jointly using the information potential of multimodal heterogeneous data. And there are some newly developed approaches have been proven to be effective in fusing multiple remote sensor data sources, although these methods got limited capabilities. To overcome this issue, the authors proposed a module called the Cross-Channel Reconstruction module to obtain a more robust and compact joint feature representation stored by the encoded latent vector. 

For previous traditional modal fusion methods, the most direct way is to concatenate different features as a fused embedding, depending on the stage in which the features are concatenated, there were early fusion, middle fusion, and late fusion. But these naive methods showed limited performance. To conquer this they have the CCR fusion which uses an encoder-decoder-like structure to obtain robust feature representation. As you can see in the detailed illustration the features extracted by the CNN are primarily simply concatenated by a prescribed order. This concatenated feature is fed into an encoder to generate a more compact feature encoding. The compressed feature encoding will be further decoded to a swapped original extracted feature. For the objective function, the holistic loss is decomposed into a cross-entropy loss term to judge classification accuracy and a reconstruction loss term to reflect how well the decoding works.

From \cite{boltzmann2012}, the generative model that can classify different types of information using multimodal Deep Boltzmann Machine.
It extracts a unified representation that fuses modalities together and learns probability density over the space of multimodal inputs.
It can extract this representation even when some modalities are absent by sampling from the conditional distribution over them and filling them in. There have been several approaches to learning from multimodal data. Huiskes showed that using captions, or tags, in addition to standard low-level image features, significantly improves the classification accuracy of Support Vector Machines (SVM) and Linear Discriminant Analysis (LDA) models. Multiple kernel learning framework by Guillaumin, further demonstrated that an additional text modality could improve the accuracy of SVMs on various object recognition tasks.

\subsection{Fall Detection}

One multimodal approach for fall detection involves combining radar and optical cameras to detect falling. \cite{zhou2018fall} proposed using convolution neural networks with multi-sensor fusion, proposed using a combination of radar and optical camera to detect falling For the radar-returned signals, the time-frequency (TF) domain is typically used. TF signals of a fall and three common motions: walking, squatting, and standing up. The authors proposed using the Alex-based convolutional neural network (CNN) and single shot multi-box detector (SSD) Net based CNN is adopted to classify the TF images respectively, and the results of the two CNNs are merged to give the final detection results. On the other hand, the image sequence captured by the optical camera is also processed by another SSD-based CNN, and the aspect ratio sequence of the bounding box of the human is provided to help confirm if a fall event is occurring. The aspect ratio changing of the bounding boxes (height divided by width) frame by frame is used to help distinguish the fall and non-fall events.

\cite{fallml2016} aimed at the recognition of and the differentiation between fall activities and activities of daily living (ADL) using the MobiFall dataset with the help of machine learning algorithms. They used an accelerometer, gyroscope, and magnetometer to collect the body movements. The five classification methods that were implemented are: Naive Bayes,k-nearest neighbor, artificial neural networks, and least square method

\begin{figure}[ht]
    \includegraphics[width = 9cm]{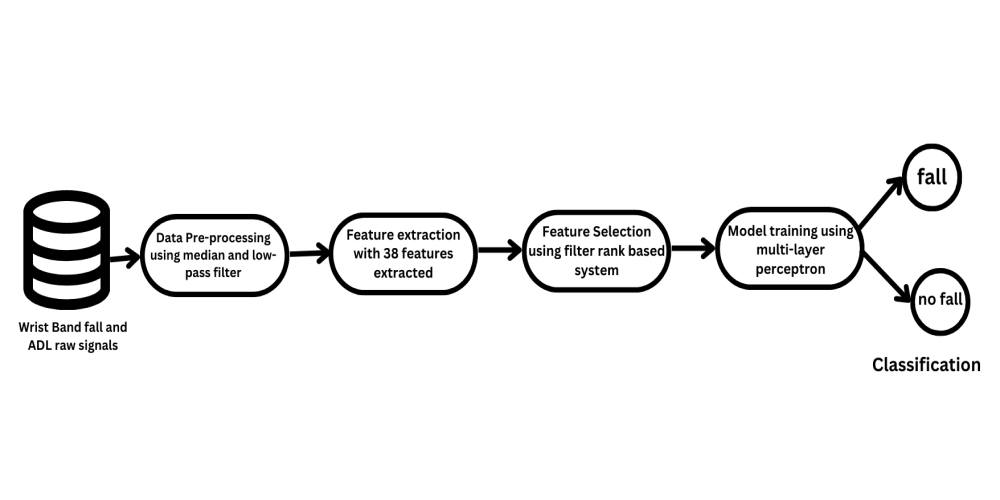}
    \caption{Fall detection approach}
    \label{fall}
\end{figure}

Fall detection classification systems were characterized through five stages: pre-processing, feature extraction, feature selection, model training, and classification. The data was pre-processed by a median and low-pass filter. During the feature extraction stage, a total of 38 features were extracted and a filter rank-based system was used to eliminate features with no information and extract the top five features in order to optimize the algorithm’s dimensionality. (z median and x median which are time…..x mean and y mean in frequency…..skewness in signal magnitude vector) can be seen in Fig. ~\ref{fall}.
Receiver Operating Characteristic curves tell that among 5 classifiers, k- Nearest Neighbors’ algorithm obtained an overall accuracy of 87.5\% with a sensitivity of 90.70\%, and a specificity of 83.78\%. Sensitivity represents the capacity to detect falls, specificity represents the capacity to only detect falls and ignore the non-fall events, and accuracy represents the portion of true results among the population.

The limitation of this paper is “MobiFall” dataset was recorded in the user’s pocket, which is below the waist but through research, it was found that the device at the upper trunk of the body, below the neck, and above the neck provides the best results. The reason being was that the other locations contain high movement frequency and complexity.

\cite{accelerometer2019} focuses on fall detection based on deep learning methods. The type of data used in this approach is accelerometer data collected from wearable devices. The model is deployed on a fog device, in this specific case, the authors use Raspberry Pi for the experiment. The authors also proposed a convolutional architecture called CNN-3B3Conv which composite of 3 blocks. Finally, this approach reaches an accuracy of 99.86\%, while the pure LSTM model only achieves around 95\%. The architecture uses 3 blocks. Block 1 and Block 2 are in the same shape with 3 convolutional layers each. For block 3 there are 3 fully-connected layers. Finally, the output is a binary label representing whether the user is falling or not.

\section{Proposed Method}
\subsection{Dataset}
\begin{figure}[b]
\centerline{\includegraphics[width=8cm]{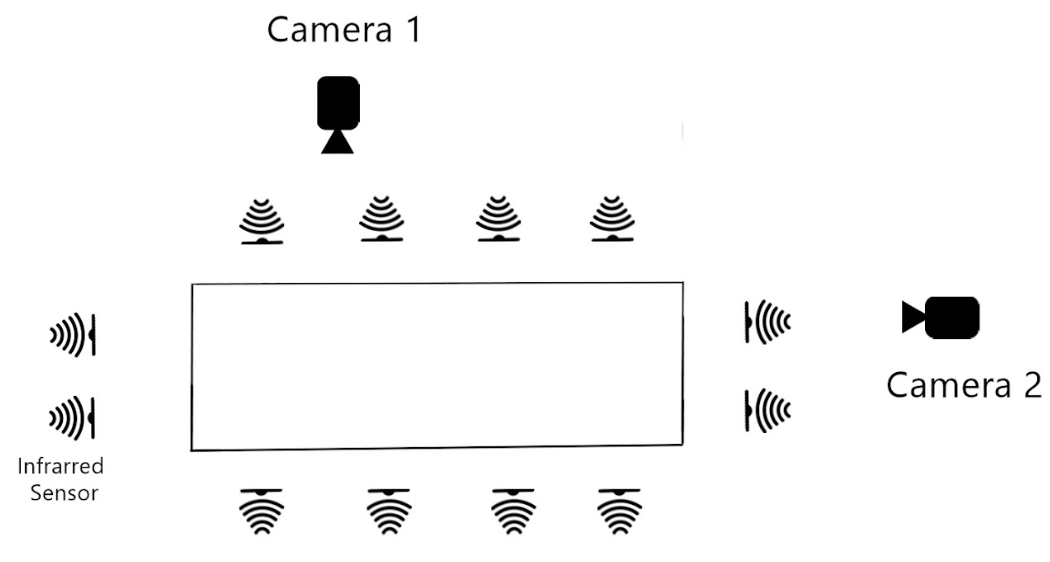}}
\caption{Layout of sensors for the experiment in UP-FALL dataset }
\label{sensor}
\end{figure}

We use the UP-FALL detection dataset ~\cite{martinez2019up} published by Panamerican University. This dataset is used for human activity recognition and is mainly aimed at detecting falls of elderly people. The dataset contains data from 17 subjects; each subject performs 11 activities with three trials for each activity. The activities were six simple human daily activities (walking, standing, picking up an object, sitting, jumping, and laying) and five human falls (falling forward using hands, falling forward using knees, falling backward, falling sitting in an empty chair, and falling sideways). Devices utilized to collect data include two cameras (one mounted in front of the subject and one mounted to the side of the subject); 12 infrared sensors surrounding the test scene as shown in Fig. ~\ref{sensor}; a brainwave sensor (electroencephalography helmet), and six wearable devices bound to different locations on the body to detect acceleration, angular velocity, and luminosity. There are two types of data in this dataset. One is the image data captured by the cameras, and the other is the sensor data stored in a CSV file. The data collected in this dataset for detecting human behavior is very comprehensive. The raw labels include not only fall or no fall but also other static behavioral labels such as standing, laying, etc.

\subsection{Preprocessing}
The main goal of this project is to binary detect falls or not. The original labels of the dataset include behavioral labels other than falling or not, which seems to be redundant at this stage. In the preprocessing, we keep the falling label unchanged and set all the remaining labels as not falling. This operation is designed to match the experimental goal of determining whether or not a fall occurs on a frame-by-frame basis. Furthermore, we used only three subjects out of the seventeen subjects. In addition, we considered in the context of real-world usage that although the data collected in the original dataset was comprehensive, it would be impractical for the elderly to wear up to six wearable sensors all over their body to detect falls. 

\begin{figure}[t]
\centerline{\includegraphics[width=5cm]{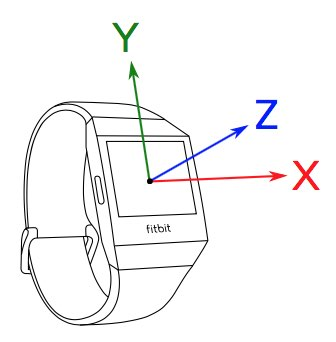}}
\caption{Example of accelerometer showing x, y, z axes with FitBit }
\label{fb}
\end{figure}

\begin{figure}[t]
\centerline{\includegraphics[width=8cm]{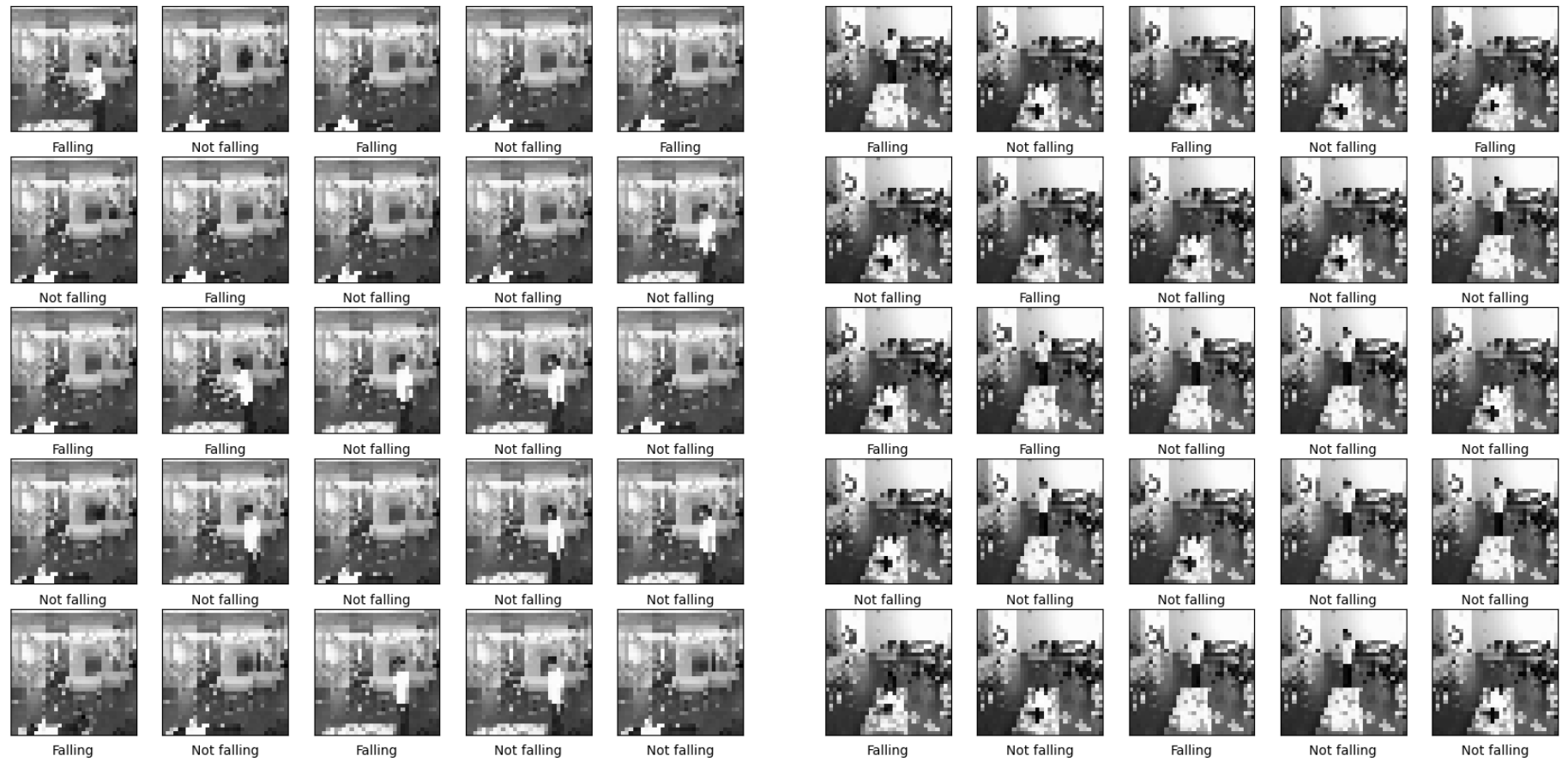}}
\caption{Preprocessed grayscaled 32 $\times$ 32 images from cameras 1 \& 2 with binary labeling of falling or not falling. }
\label{dv}
\end{figure}

\begin{figure*}[ht!]
\centerline{\includegraphics[width=14cm]{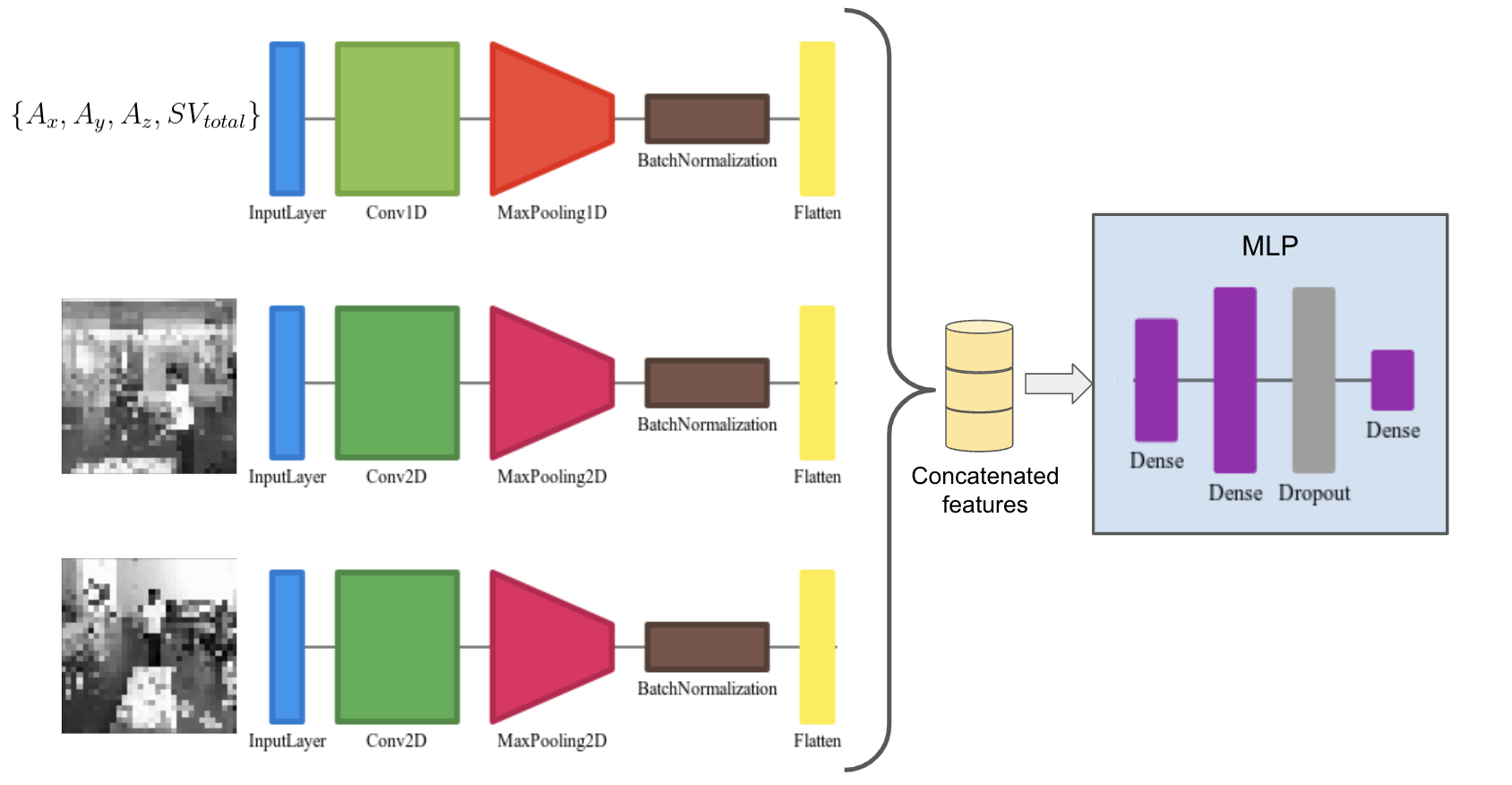}}
\caption{Muitimodality approach.}
\label{multim}
\end{figure*}

Thus we reduced the number of sensors from six to one, keeping only the accelerometer on the wrist because the implementation of embedding sensors into smartwatches is well established and fully commercialized, and wearing a wristwatch-type device will have minimal impact on the wearers' daily life shown in Fig. \ref{fb}. We also decided not to use angular velocity and luminosity data for the sensor, keeping only the accelerometer data. Raw accelerometer data were collected in the x, y, and z directions, forming a three-dimensional vector. We calculate the magnitudes of the raw accelerometer data by the following formula,
\begin{equation}
    SV_{total} = \sqrt{A_{x}^2+A_{y}^2+A_{z}^2}
\end{equation}
where $A_{x}$, $A_{y}$, and $A_{z}$ denote raw accelerometer data along three axes, and $SV_{total}$ denotes the computed magnitudes.
The rest of the sensors (brainwave sensor and infrared sensors) were also eradicated, as it is not practical to deploy specialized equipment in a practical scenario. Finally, we obtained the preprocessed sensor data, which are the accelerations in x, y, and z directions collected by the wrist-bounded accelerometers and their magnitudes.\par
The images (frames) are converted to grayscale and resized to 32 $\times$ 32. A visualization of preprocessed images with corrected labels annotated can be seen in Fig. ~\ref{dv}.
\begin{figure}[ht!]
        \centering
        \begin{subfigure}[b]{0.3\textwidth}
            \centering
            \includegraphics[width=\textwidth]{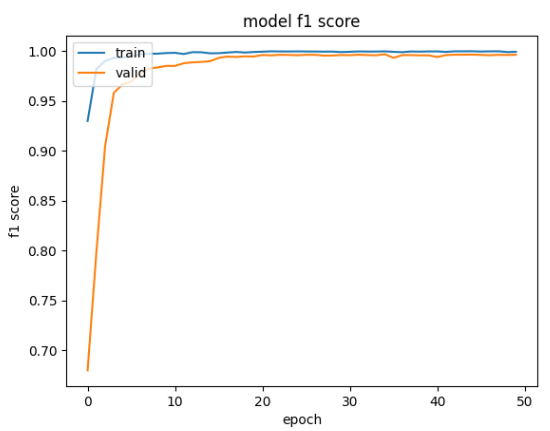}
            \caption{Evaluation of Mono-Sensor, Wrist Accelerometer, of F1 Performance over 50 epochs}   
            \label{monos}
        \end{subfigure}
        \begin{subfigure}[b]{0.3\textwidth}  
            \centering 
            \includegraphics[width=\textwidth]{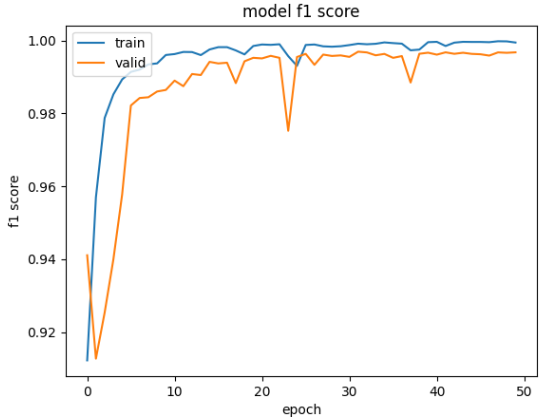}
            \caption{Evaluation of Multi-Sensor of F1 Performance over 50 epochs}   
            \label{multis}
        \end{subfigure}
        \caption{Visualization of F1 scores for mono-sensor and multi-sensors} 
        \label{monovsmulti}
    \end{figure}
\begin{figure*}[p]
        \centering
        \begin{subfigure}[b]{0.4\textwidth}
            \centering
            \includegraphics[width=\textwidth]{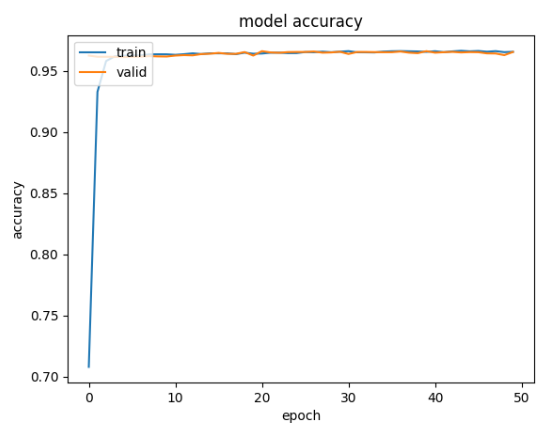}
            \caption{Wrist Sensor Accelerometer Data Unimodality Accuracy Performance over 50 epochs}   
            \label{sensora}
        \end{subfigure}
        \begin{subfigure}[b]{0.4\textwidth}  
            \centering 
            \includegraphics[width=\textwidth]{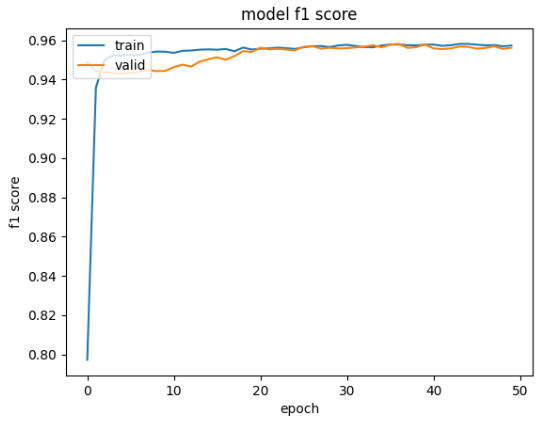}
            \caption{Wrist Sensor Accelerometer Data Unimodality F1 score Performance over 50 epochs}   
            \label{sensorf1}
        \end{subfigure}
        \vskip\baselineskip
        \begin{subfigure}[b]{0.4\textwidth}   
            \centering 
            \includegraphics[width=\textwidth]{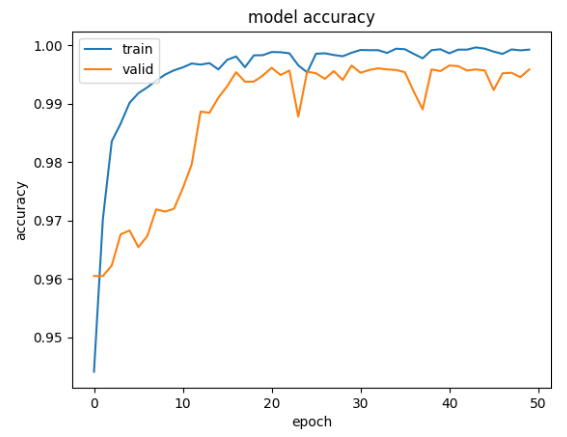}
            \caption{Camera 1 and 2 Data Unimodality Accuracy Performance over 50 epochs}   
            \label{cama}
        \end{subfigure}
        \begin{subfigure}[b]{0.4\textwidth}   
            \centering 
            \includegraphics[width=\textwidth]{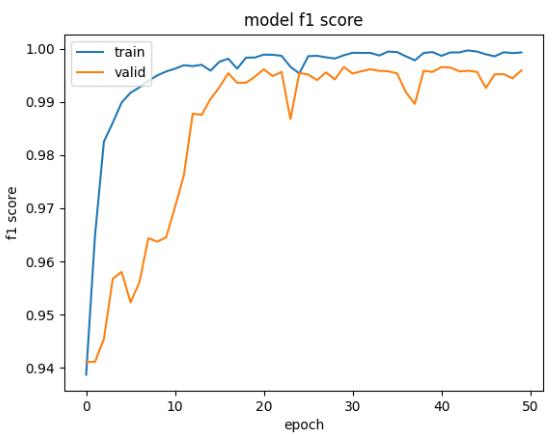}
            \caption{Camera 1 and 2 Data Unimodality F1 score Performance over 50 epochs}  
            \label{camf1}
        \end{subfigure}
        \vskip\baselineskip
        \begin{subfigure}[b]{0.4\textwidth}   
            \centering 
            \includegraphics[width=\textwidth]{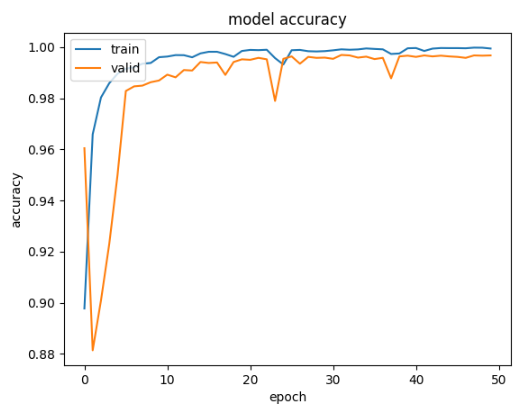}
            \caption{Multimodality Fusion Accuracy Performance over 50 epochs}  
            \label{multia}
        \end{subfigure}
        \begin{subfigure}[b]{0.4\textwidth}   
            \centering 
            \includegraphics[width=\textwidth]{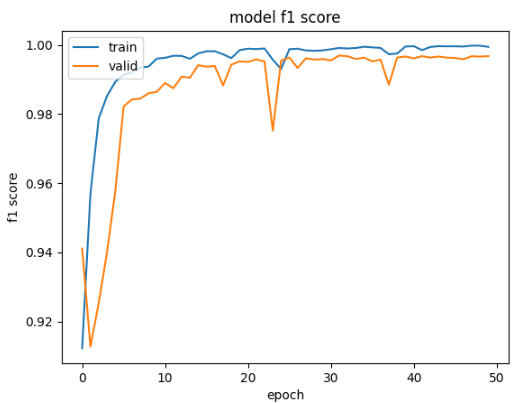}
            \caption{Multimodality Fusion F1 score Performance over 50 epochs}   
            \label{multif1}
        \end{subfigure}
        \caption{Visualization of selected metrics for unimodality and multimodality.} 
        \label{metrics}
    \end{figure*}

\subsection{Multimodality approach}
\begin{table*}[t]
    \centering
\caption{Evaluation metrics of Sensor, Camera 1, Camera 2, fusion of Camera 1 and 2, and Multimodality fusion}
\label{evaluation}
\begin{tabular}{cccccc}
    \toprule
& Sensor & Camera 1 & Camera 2 & Camera 1 \& 2 & Multimodality fusion\\
    \midrule
Accuracy & 96.47 & 99.26 & 99.45 &  99.60 & \textbf{99.72}\\
Precision & 96.47 & 99.26 & 99.45 & 99.60 &\textbf{99.72}\\
Recall & 96.47 & 99.26 & 99.45 & 99.60 &\textbf{99.72}\\
F1 score & 95.62 & 99.24 & 99.45 & 99.59 &\textbf{99.72}\\
\bottomrule
\end{tabular}
\vspace{-0.3cm}
\end{table*}
We chose an intuitive and concise method of multimodal feature fusion: directly concatenating two different sets of features. We first extract the feature maps of the image data and sensor data using convolutional neural networks and then flatten the different modal features to concatenate them. The concatenated feature embeddings are fed into a multi-layer perceptron consisting (MLP) constituted of fully-connection layers. The final layer is activated by a softmax function to output the corresponding prediction.
    
\subsubsection{Sensor data}
For sensor data, the wrist accelerometer, even though for each sample (timestamp), there are four data, i.e., $A_{x}$, $A_{y}$, $A_{z}$ and $SV_{total}$, the dimensionality of the information is mono. We first tried feature extraction using an MLP for unimodal prediction of falling or not. In addition, convolutional neural networks have good feature extraction capability. A 1D convolutional neural network can be used for one-dimensional data. Thus we also fed the sensor data into a 1D convolutional neural network for unimodal prediction. The results show that the 1D convolutional neural network performs better. And since we will apply a 2D convolutional neural network to the image data, it is obviously better for feature fusion with the same operation criterion for the sensor data.

\subsubsection{Image data}
For feature extraction of image data, the usage of convolutional networks has been proven to be superior by various studies and applications. Therefore, we use a 2D convolutional neural network for feature extraction of image data. For each frame in the video stream, we convert the RGB image into a single-channel grayscale map and then resize it to a 32 $\times$ 32 size for inputting into the convolutional neural network.

\subsubsection{Feature fusion}
For the sensor data, we feed it into a convolutional block consisting of a 1D convolutional layer, a 1D max-pooling layer, and a batch normalization layer, then flatten it into a 1D embedding. For two sets of image data, we first input a convolutional block consisting of 2D convolutional layers, 2D max-pooling layers, and batch normalization layers and then flatten it into a 1D embedding. For three sets of flattened feature embeddings, we concatenate them together and input them into an MLP consisting of two fully-connection layers and a dropout layer evading overfitting. A final fully-connection layer reduces the embedding dimension to two, and the two-dimensional embedding will be activated by the Softmax function. The obtained probabilistic scores are converted to output predictions.

\section{Evaluation}
\subsection{Classification capability}
We evaluated the results for unimodal sensor data, unimodal image data, and multimodal fusion approach, respectively. The unimodal image data already fused the extracted features from the video streams of camera one and camera 2. As evaluation metrics, we selected the accuracy and F1 scores. Accuracy is the most basic metric for a classification task, while the F1 score combines recall and precision for a more comprehensive and intuitive evaluation of a classification model. Accuracy, precision, recall, and F1 score can be calculated by the following equations, 
\begin{equation}
    accuracy = \frac{TP+TN}{TP+FN+TN+FP}
\end{equation}

\begin{equation}
    precision = \frac{TP}{TP+FP}
\end{equation}

\begin{equation}
    recall = \frac{TP}{TP+FN}
\end{equation}

\begin{equation}
    F1 score = \frac{2*precision*recall}{precision+recall}\textbf{}
\end{equation}
where $TP$, $FP$, $TN$, and $FN$ denote true positive, false positive, true negative, and false negative, respectively. The visualization of the evaluation metrics can be observed in Fig.~\ref{metrics}. Detailed numeric comparison can be seen in Table.~\ref{evaluation}.

\begin{table}[b]
    \centering
\caption{Evaluation metrics of mono-sensor (wrist) v.s multi-sensor}
\label{monovmulti}
\begin{tabular}{ccccc}
    \toprule
& Accuracy & Precision & Recall & F1 score \\
    \midrule
Mono-sensor (wrist) & \textbf{99.72} & \textbf{99.72} & \textbf{99.72} &  \textbf{99.72} \\
Multi-sensor & 99.68 & 99.68 & 99.68 & 99.68 \\

\bottomrule
\end{tabular}
\vspace{-0.3cm}
\end{table}

\subsection{Feasibility of data cleansing}
Since we reduced the number of sensors during the data preprocessing period, we experimented with training the binary prediction model using the original sensor data after implementing our method to ensure that the cleansing operation did not impact the model's binary prediction performance for fall detection. The F1 scores of the two methods are visualized in Fig.~\ref{monovsmulti}, and it can be seen that there is almost no difference in the best F1 scores of the two methods on the validation set. Although from the images, the model using multiple sensors performs more consistently and converges faster during training, the single-sensor model oscillates more. To further verify the feasibility of the data refinement operation, we further evaluated it on the test set. Numeric results can be seen in Table ~\ref{monovmulti}. On the test set, our model even shows a slight advantage. We can conclude that the cleansing of the sensor data did not affect the model's classification performance.

\section{Discussion and Conclusion}
\subsection{Conclusion}
To conclude,we cleaned the data set reducing seventeen subjects to three with eleven activities and three trials. Six sensors were reduced to use only wrist sensor and we excluded angular velocity and luminosity to use only accelerometer for the prediction. All of the labels were reduced to only two, namely fall and no fall. For multi modal approach, we concatenated two different set of features for fall detection. Furthermore, we used fusion of sensor as well as image data to increase the performance. This data was fed to multi layer perceptron having fully connected layers and final layer was activated using the softmax function. By using all these experiments, we achieved an accuracy,precision, recall and F1 score of 99.72 percent as comapred to 99.68 percent for multi sensor data.

\subsection{Future Works}
There are many options for future work we could do to improve this project. For example, in the future, we would aim to focus more on feature extraction to get more understanding related to falling. We would focus on experimenting with RNN or LSTM neural networks to improve the performance of fall detection problems. In this, we are aiming to split frames into time windows and perform time series prediction. Along with that, we will be expanding feature dimensions after achieving time windows. We realised that even if the accuracy did not change with using only wrist data as compared combining all the data, but we focus to try different combination of categories to increase performance of model in future.

\bibliographystyle{IEEEtran}
\bibliography{bib.bib}

\end{document}